\documentclass{bmcart}

\usepackage{amsthm}
\usepackage{amsmath}
\usepackage{amssymb}
\usepackage{array}
\usepackage[]{algorithm2e}
\usepackage[T1]{fontenc}

\usepackage{graphicx}
\usepackage{tabularx}
\providecommand{\tabularnewline}{\\}

\usepackage{multirow}

\startlocaldefs
\endlocaldefs

\begin{document}

\begin{frontmatter}

\begin{fmbox}
\dochead{Research}


\title{Mining Massive Hierarchical Data Using a Scalable Probabilistic Graphical Model}


\author[
   addressref={aff1},                   
   corref={aff1},                       
   email={aljadda@uga.edu}   
]{\inits{KA}\fnm{Khalifeh} \snm{AlJadda}}
\author[
   addressref={aff2},                   
   email={mkorayem@cs.indiana.edu}   
]{\inits{MK}\fnm{Mohammed} \snm{Korayem}}
\author[
   addressref={aff3},                   
   email={camilo.ortiz@careerbuilder.com}   
]{\inits{CO}\fnm{Camilo} \snm{Ortiz}}
\author[
   addressref={aff3},                   
   email={trey.grainger@careerbuilder.com}   
]{\inits{TG}\fnm{Trey} \snm{Grainger}}
\author[
   addressref={aff1},
   email={jam@cs.uga.edu}
]{\inits{JAM}\fnm{John A.} \snm{Miller}}
\author[
   addressref={aff1},
   email={khaled@cs.uga.edu}
]{\inits{KR}\fnm{Khaled} \snm{Rasheed}}
\author[
   addressref={aff1},
   email={kochut@cs.uga.edu}
]{\inits{KK}\fnm{Krys J.} \snm{Kochut}}
\author[
   addressref={aff4},
   email={will@ccrc.uga.edu}
]{\inits{WY}\fnm{William S.} \snm{York}}
\author[
   addressref={aff4},
   email={rene@ccrc.uga.edu}
]{\inits{RR}\fnm{Rene} \snm{Ranzinger}}
\author[
   addressref={aff4},
   email={mindyp@ccrc.uga.edu}
]{\inits{MP}\fnm{Melody} \snm{Porterfield}}


\address[id=aff1]{
  \orgname{Department of Computer Science, University of Georgia}, 
  \city{Athens,GA},                              
  \cny{USA}                                    
}
\address[id=aff2]{%
  \orgname{School of Informatics and Computing, Indiana University},
  \city{Bloomington, IN},
  \cny{USA}
}
\address[id=aff3]{%
  \orgname{CareerBuilder.com},
  \city{Norcross, GA},
  \cny{USA}
}
\address[id=aff4]{
  \orgname{Complex Carbohydrate Research Center, University of Georgia}, 
  \city{Athens,GA},                              
  \cny{USA}                                    
}

\begin{artnotes}
\note[id=n1]{Equal contributor} 
\end{artnotes}

\end{fmbox}


\begin{abstractbox}

\begin{abstract} 
Probabilistic Graphical Models (PGM) are very useful in the fields of machine learning and data mining. The crucial limitation of those models,however, is the scalability. The Bayesian Network, which is one of the most common PGMs used in machine learning and data mining, demonstrates this limitation when the training data consists of random variables, each of them has a large set of possible values. In the big data era, one would expect new extensions to the existing PGMs to handle the massive amount of data produced these days by computers, sensors and other electronic devices. With hierarchical data - data that is arranged in a treelike structure with several levels - one would expect to see hundreds of thousands or millions of values distributed over even just a small number of levels. When modeling this kind of hierarchical data across large data sets, Bayesian Networks become infeasible for representing the probability distributions. 
In this paper we introduce an extension to Bayesian Networks to handle massive sets of hierarchical data in a reasonable amount of time and space. The proposed model achieves perfect precision of $1.0$ and high recall of $0.93$ when it is used as multi-label classifier for the annotation of mass spectrometry data. On another data set of 1.5 billion search logs provided by CareerBuilder.com the model was able to predict latent semantic relationships between search keywords with accuracy up to 0.80.

\end{abstract}



\end{abstractbox}
%

\end{frontmatter}



\section*{Introduction}
Probabilistic graphical models (PGM) consist of a structural model and a set of conditional probabilities \cite{1,2}. They are widely used in machine learning and data mining techniques, like classification, speech recognition~\cite{korayem2007optimizing}, bioinformatics~\cite{eddy1994rna,soding2005protein}, Natural Language Processing (NLP) \cite{christodoulopoulos2011bayesian,kupiec1992robust}, etc. Scalability and restricted domain size (e.g., propositional domain) are the major challenges for PGMs. To overcome these challenges one would expect extensions to the existing PGMs. One extension is offered by the hierarchical probabilistic graphical models (HPGM) which aim to extend the PGM to work with more structured domains \cite{15,fine1998hierarchical}. However, this extension tackles the restricted domain size problem, but not scalability. 
For hierarchical data, where data can be divided into several levels arranged in tree-like structures, data items in each level depend on or are influenced only by the data items in the upper levels while a Bayesian Network (BN) is the most appropriate PGM to represent a probability distribution since the dependencies in this kind of data are not bidirectional, a BN is often infeasible, as it may not provide a concise enough representation of a large probability distribution.When dealing with the kind of massive hierarchical data that is becoming increasingly common in the big data era, this is true because each level represents a random variable, while each node in that level represents an outcome (possible value) of that random variable, so the data can grow horizontally (number of values) faster than vertically (number of random variables). Moreover, since the dependency between the random variables is pre-defined in the hierarchical data, the structure of the network is predefined. Hence, the first phase of building Bayesian Network to find the optimal structure is not applicable

For example, consider the glycan ontology "GlycO" \cite{thomas2006modular} which describes 1300 glycan structures (see section \ref{sec:Exp}) whose theoretical tandem mass spectra (MS) can be predicted by GlycoWorkbench \cite{24}. If the maximum of cleavages is set to two and the number of cross-ring cleavages is set to one,  then the theoretical MS$^2$ spectrum contains 2,979,334 ions, which themselves can be fragmented to form tens of millions of ions in MS$^3$. To represent this data set of only two levels of the MS data using a Bayesian Network (BN) the network will be composed of two nodes, MS$^1$ and MS$^2$ with a single path $MS^1 \rightarrow MS^2$ while the conditional probability table (CPT) for the $MS^2$ will contain 3,873,134,200 (2,979,334 $\times$ 1300) entries. For this kind of data, we propose a simple probabilistic graphical model for massive hierarchical data (PGMHD) which we consider as an extension to the Bayesian Network (BN,) that can represent massive hierarchical data in a more efficient way. We successfully apply the PGMHD in two different domains: bioinformatics (for multi-label classification) and search log analytics (for latent semantic discovery of related terms, as well as, semantically ambiguous terms).

The main contributions of this paper are as follows:
We propose a simple, efficient and scalable probabilistic model that extends Bayesian Network for massive hierarchical data.
We successfully apply this model to the bioinformatics domain in which we automatically classify and annotate high-throughput mass spectrometry data.
We also apply this model for large-scale latent semantic
discovery and semantically ambiguous terms discovery using 1.6 billion search log entries provided by CareerBuilder.com, using the Hadoop Map/Reduce framework.

\section*{Background}
Graphical models can be classified into two major categories: (1) directed graphical models (the focus of this paper), which are often referred to as Bayesian Networks, or belief networks, and (2) undirected graphical models which are often referred to as Markov Random Fields, Markov networks, Boltzmann machines, or log-linear models \cite{3}. Probabilistic graphical models (PGMs) consist of both graph structure and parameters. The graph structure represents a set of conditionally independent relations for the probability model, while the parameters consist of the joint probability distributions \cite{1}.
Probabilistic graphical models are often considered to be more convenient than numerical representations for two main reasons \cite{4}:
\begin{enumerate}
\item To encode a joint probability distribution for P($X_1$,...,$X_n$) for $n$ propositional random variables with a numerical representation, we need a table with $2^n$ entries.
\item Inadequacy in addressing the notion of independence: to test independence between $X$ and $Y$, one needs to test whether the joint distribution of $x$ and $y$ is equal to the product of their marginal probability.
\end{enumerate}

PGMs are used in many domains. For example, Hidden Markov Models (HMM) are considered a crucial component for most of the speech recognition systems~\cite{korayem2007optimizing}. In bioinformatics, probabilistic graphical models are used in RNA sequence analysis~\cite{eddy1994rna}. In natural language processing (NLP), HMM and Bayesian models are used for part of speech (POS) tagging~\cite{christodoulopoulos2011bayesian}. The problem with PGMs in general, and Bayesian Networks in particular, is that they are not suitable for representing massive data due to the time complexity of learning the structure of the network and the space complexity of storing a network with thousands of random variables or random variables taking in many values. In general, finding a network that maximizes the Bayesian score which maximizes the posterior probability and Minimum Description Length (MDL) score which gives preference to a simple BN over a complex one, is an NP-hard problem \cite{5}.

\subsection*{Bayesian Network}
\begin{figure}
\centering
\includegraphics[scale = 0.7]{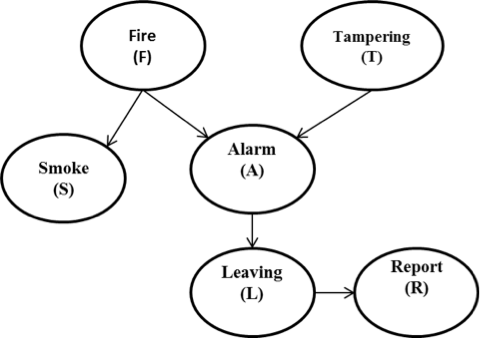}
\caption{Bayesian Network}
\label{bn}
\end{figure}

A Bayesian Network is a concise representation of a large probability distribution to be handled using traditional techniques such as tables and equations \cite{6}. The graph of a Bayesian Network is a directed acyclic graph (DAG)~\cite{2}. A Bayesian Network consists of two components: a DAG representing the structure (as shown in Figure \ref{bn}), and a set of conditional probability tables (CPTs). Each node in a Bayesian Network must have a CPT which quantifies the relationship between the variable represented by that node and its parents in the network. Completeness and consistency are guaranteed in a Bayesian Network since there is only one probability distribution that satisfies the Bayesian Network constraints~\cite{6}.
The constraints that guarantee a unique probability distribution are the numerical constraints represented by CPT and the independence constraints represented by the structure itself. The independence constraints is shown in Figure \ref{bn}. Each variable in the structure is independent of any other variables other than its parents, once its parents are known. For example, once the information about A is known, the probability of L will not be affected by any new information about F or T, so we call L independent of F and T once A is known.

Bayesian Networks are widely used for modeling causality in a formal way, for decision-making under uncertainty, and for many other applications \cite{6}.

\section*{Related Work}
Our research is related closely to Bayesian Network classifiers. In this section we review different forms of Bayesian Network classifiers to understand how the PGMHD extends BN in a way different than the existing models. We will cover the following BN classifiers:
\begin{enumerate}
\item Naive Bayes Classifier (NB).
\item Selective Naive Bayes (SNB)
\item Tree Augmented Naive Bayes (TAN).
\item Hidden Naive Bayes (HNB).
\end{enumerate}

we will also cover how we applied PGMHD to other data mining problems, such as, latent semantic discovery of related search terms in users search logs, and of semantically ambiguous keywords by analyzing users' search logs.

\subsection*{Naive Bayes (NB)}
\begin{figure}
\centering
\includegraphics[scale=0.5]{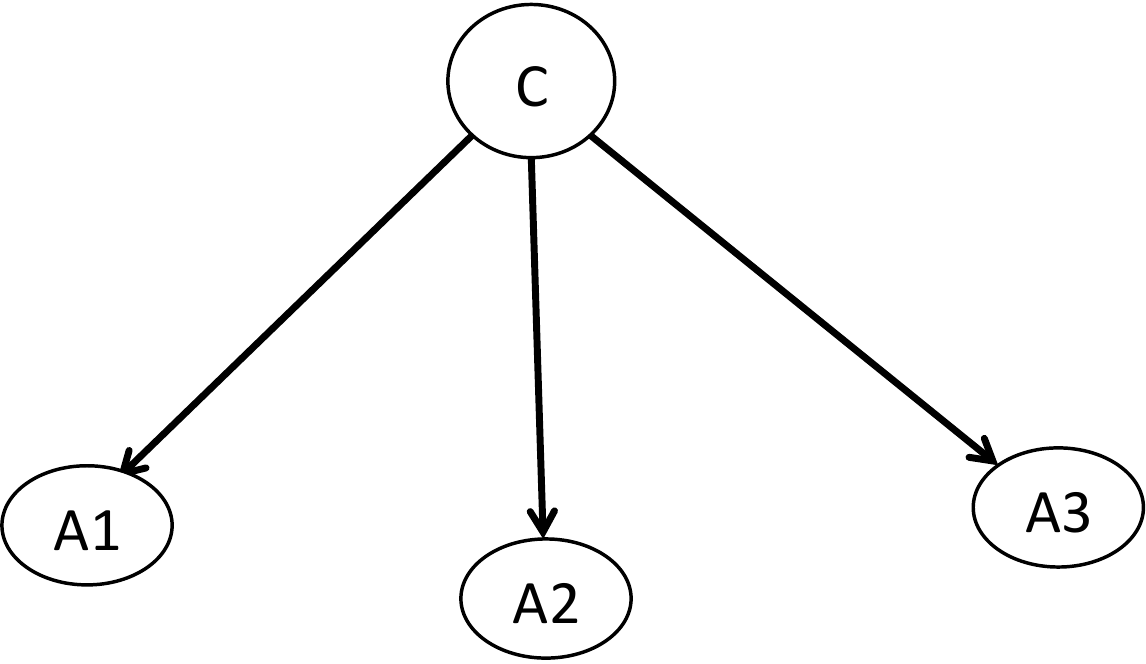}
\begin{center}
\caption{Naive Bayes}
\label{nb}
\end{center}
\end{figure}

Naive Bayes is the simplist form of the BN classifiers and the most common one. This classifier is based on an assumption that all the features are independent given the class. Figure \ref{nb} shows an example of NB. A NB classifier is defined as follows:
\[
P(c|{\bf x})\propto P(c) \prod_{j=1}^n P(x_j|c) 
\]
Where ${\bf x} = (x_1,..,x_n)$. $P(c)$ is the prior probability of class $c$ and $P(x_j|c)$ is the conditional probability of feature/variable $x_j$. The value of $c$ that maximizes the right hand side is chosen.
A Naive Bayes classifier's performance depends upon the quality of the predictor features, such that the performance is improved when the predictor features are relevant and non redundant.

\subsection*{Selective Naive Bayes (SNB)}
In order to improve the performance of BN classifier by selecting the predictive features that are relevant and not redundant, Selective Naive Bayes (SNB) \cite{bielza2014discrete} is proposed as a feature subset selection problem. Let us define ${\bf x}_F$ as the projection of ${\bf x}$ onto a selected feature subset ${\bf F} \subset {\{1,2,\ldots,n\}}$. The classification equation becomes

\[
P(c|{\bf x})\propto P(c|{\bf x}_F) = P(c) \prod_{j\in F}P(x_j|c) 
\]

\subsection*{Tree Augmented Naive Bayes (TAN)}
\begin{figure}
\centering
\includegraphics[scale=0.5]{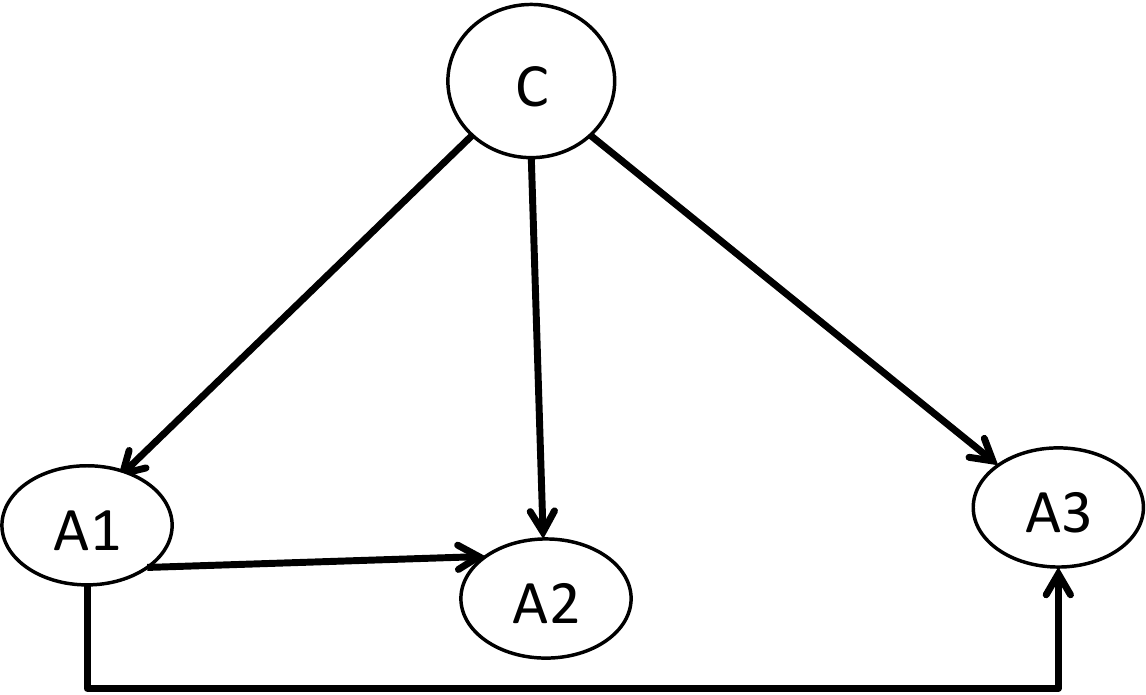}
\begin{center}
\caption{Tree Augmented Naive Bayes}
\label{tan}
\end{center}
\end{figure}

This form of Bayesian Network classifier extends the NB by allowing each attribute to have at most one attribute parent in addition to its class parent. This extension tends to represent the fact that in some cases there is dependency or influence between features in a way that a value of a feature $x_j$ depends on a value of feature $y$. TAN classifier is defined as follows:
\[
P(c|{\bf x}) = P(c) \prod_{j=1}^n P(x_j|p_{xj},c) 
\]
Where $p_{xj}$ is the attribute parent of $x_j$. Figure \ref{tan} shows an example of TAN.

\subsection*{Hidden Naive Bayes (HNB)}
\begin{figure}
\centering
\includegraphics[scale=0.5]{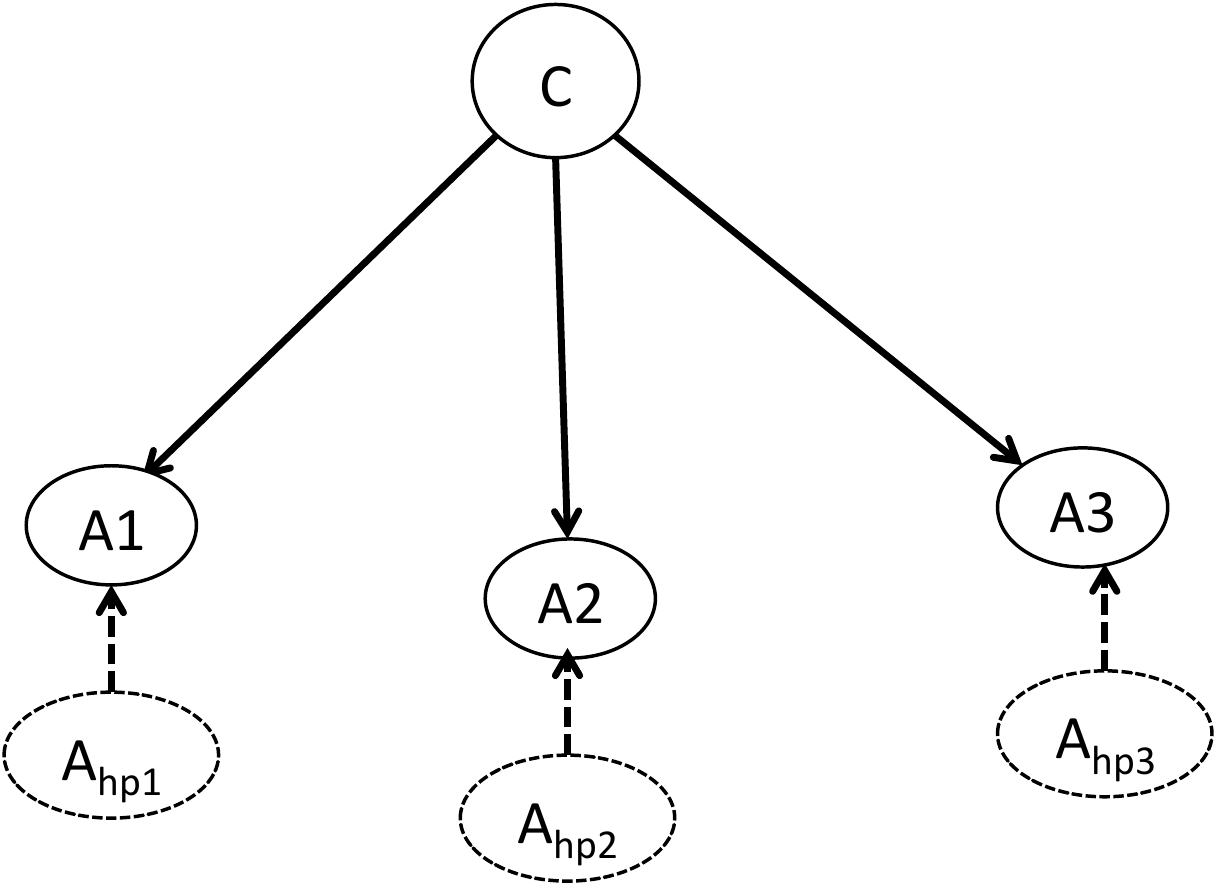}
\begin{center}
\caption{Hidden Naive Bayes}
\label{hnb}
\end{center}
\end{figure}

HNB (Figure \ref{hnb}) is another extension of NB. In this extension each attribute $A_i$ gets a hidden parent $A_{hpi}$ to integrate the influences from all other attributes. The definition of the HNB classifier is as follows:
\[
P(c|{\bf x}) = P(c) \prod_{j=1}^n P(x_j|h{p_j},c) 
\]
where,
\[
P(x_j|x_{hpj},c)  = \sum_{i=1,i\neq j}^n w_{ji} * P(x_j|x_i,c) 
\]

\section*{Probabilistic Graphical model for Massive Hierarchical Data Problems (PGMHD)}
In this section we describe PGMHD. We discuss the structure of the model, its learning algorithm, and how it extends BN.

\subsection*{Model Structure}
Consider a multi-level directed graph $G=(V,A)$ where $V$ and $A\subset V\times V$
denote the sets of nodes and arcs, respectively, such that:
\begin{enumerate}
\item  $V$ is partitioned into $m$ levels $L_{0},\ldots,L_{m-1}$
such that $V=\cup_{i=0}^{m-1}L_{i}$, and $L_{i}\cup L_{j}=\emptyset$
for $i\not=j$.
\item The arcs in $A$ only connect one level to the next, i.e., if $a\in A$
then $a\in L_{i-1}\times L_{i}$ for some $i=1,\ldots,m-1$.
\item An arc $a=(v_{i-1},v_{i})\in L_{i-1}\times L_{i}$ represents the
dependency of $v_{i}$ with its parent $v_{i-1}$, $i=1,\ldots,m-1$.
Moreover, let $\mbox{pa}:V\to\mathcal{P}(V)$ be the function that maps every node to its parents, i.e.,
\[
\mbox{pa}(v)=\{w:(w,v)\in A\}\qquad \forall v\in V.
\]

\item The nodes in each level $L_{i}$ represent all the possible outcomes
of a finite discrete random variable, namely $X_{i}$, $i=1,\ldots,m-1$.
\end{enumerate}
Note that the nodes in the 
first level $L_{0}$ can be seen as root nodes
and the ones in $L_{m-1}$ as leaves. Also, an observation $x$ in
our probabilistic model is an outcome of a random variable, namely
$X\in L_{0}\times\cdots\times L_{m-1}$, defined as 
\begin{equation}
X=(X_{0},X_{1},\ldots,X_{m-1}),\label{randX}
\end{equation}
which represents a path from $L_{0}$ to $L_{m-1}$ such
that $(X_{i-1},X_{i})\in A$.

In addition, we assume that there are $t$ observations of $X$, namely $x^{1},\ldots,x^{t}$,
and let $f:V\times V\to\mathbb{N}$ be a frequency function defined
as 
$f(w,v)=$ Frequency of Co-Occurrence $w$ and $v$.
 Moreover, these latter $t$ observations are the ones used to train our model, so that $f(w,v)>0$ for every $(w,v)\in A$.

It should be observed that the proposed model can be seen as a special
case of a Bayesian Network by considering a network consisting of
directed predefined paths. However, we believe that a
leveled directed graph that explicitly defines one node per outcome of the random variables (as described above): i) leads to an
easily scalable (and distributable) implementation of the problems
we consider; ii) improves the readability and expressiveness of the
implemented network; and iii) simplifies and facilitates the training
of the model.

\subsection*{Probabilistic-based Classification}


Given an outcome at
level $i\in\{1,\ldots, m-1\}$, namely $v\in L_{i}$, we calculate the \emph{classification
score} $\mbox{Cl}_{i}(w|v)$ of $v$ to the parent outcome
$w\in L_{i-1}$ by estimating the conditional probability $P(X_{i-1}=w|X_{i}=v)$
as follows 
\begin{eqnarray*}
\mbox{Cl}_{i}(w|v) & := & \frac{f(w,v)}{\mbox{In}(v)}=\frac{\left(\frac{f(w,v)}{\mbox{Out}(w)}\right)\cdot\left(\frac{\mbox{Out}(w)}{t}\right)}{\left(\frac{\mbox{In}(v)}{t}\right)}\\
 & = for & \frac{P(X_{i}=v|X_{i-1}=w)\cdot P(X_{i-1}=w)}{P(X_{i}=v)}\\
 &=&P(X_{i-1}=w|X_{i}=v),
\end{eqnarray*}
where 
\[
\mbox{In}(v):=\sum_{u\in\mbox{pa}(v)}f(u,v),\qquad\forall v\in V,
\]
and
\[
\mbox{Out}(w):=\sum_{u:(w,u)\in A}f(w,u), \qquad\forall w\in V.
\]
\subsection*{Probabilistic-based Similarity scoring}
Fix a level $i\in\{1,\ldots, m-1\}$, and let $X,Y\in {L_{0}\times\cdots\times L_{m-1}}$  be identically
distributed random variables as in \eqref{randX}. We define the \emph{probabilistic-based similarity score CO (Co-Occurrence)} 
between two outcomes $x_{i},y_{i}\in L_{i}$ by computing the
conditional joint probability \[\mbox{CO}(x_{i},y_{i}):=P(X_{i}=x_{i},Y_{i}=y_{i}|X_{i-1}\in\mbox{pa}(x_{i})\cap\mbox{pa}(y_{i}),Y_{i-1}\in\mbox{pa}(x_{i})\cap\mbox{pa}(y_{i}))\]
as
\[
\mbox{CO}(x_{i},y_{i})=\prod_{w\in\mbox{pa}(x_{i})\cap\mbox{pa}(y_{i})}p_{i}(w,x_{i})\cdot\prod_{w\in\mbox{pa}(x_{i})\cap\mbox{pa}(y_{i})}p_{i}(w,y_{i}),
\]
where $p_{i}(w,v)=P(X_{i-1}=w,X_{i}=v)$ for every $(w,v)\in L_{i-1}\times L_{i}$.
We can naturally estimate the probabilities $p_{i}(v,w)$ with $\hat{p}(v,w)$
defined as
\[
\hat{p}(w,v):=\frac{f(w,v)}{Out(w)}.
\]
Hence, we can obtain the related outcomes of $x_{i}\in L_{i}$ (at
level $i$) by finding all the $y\in L_{i}$ with a large estimated\emph{
}probabilistic similarity score $\mbox{CO}(x_{i},y_{i})$.

\subsection*{Progressive Learning}
PGMHD is designed to allow progressive learning which is shown in Algorithm \ref{alg:learn}. Progressive learning is a learning technique that allows a model to learn gradually over time. Training data does not need to be given at one time to the model. Instead, the model can learn from any available data and integrate the new knowledge incrementally. This learning technique is very attractive in the big data age for the following reasons:
\begin{enumerate}
\item Training the model does not require processing all data upfront
\item It can easily learn from new data without the need to re-include the previous training data in the learning.
\item The training session can be distributed instead of doing it in one long-running session.
\item It supports recursive learning which allows the results of the model to be used as new training data, provided they are judged to be accurate by the user. 
\end{enumerate}  

\begin{figure}
\centering
\includegraphics[scale=0.65]{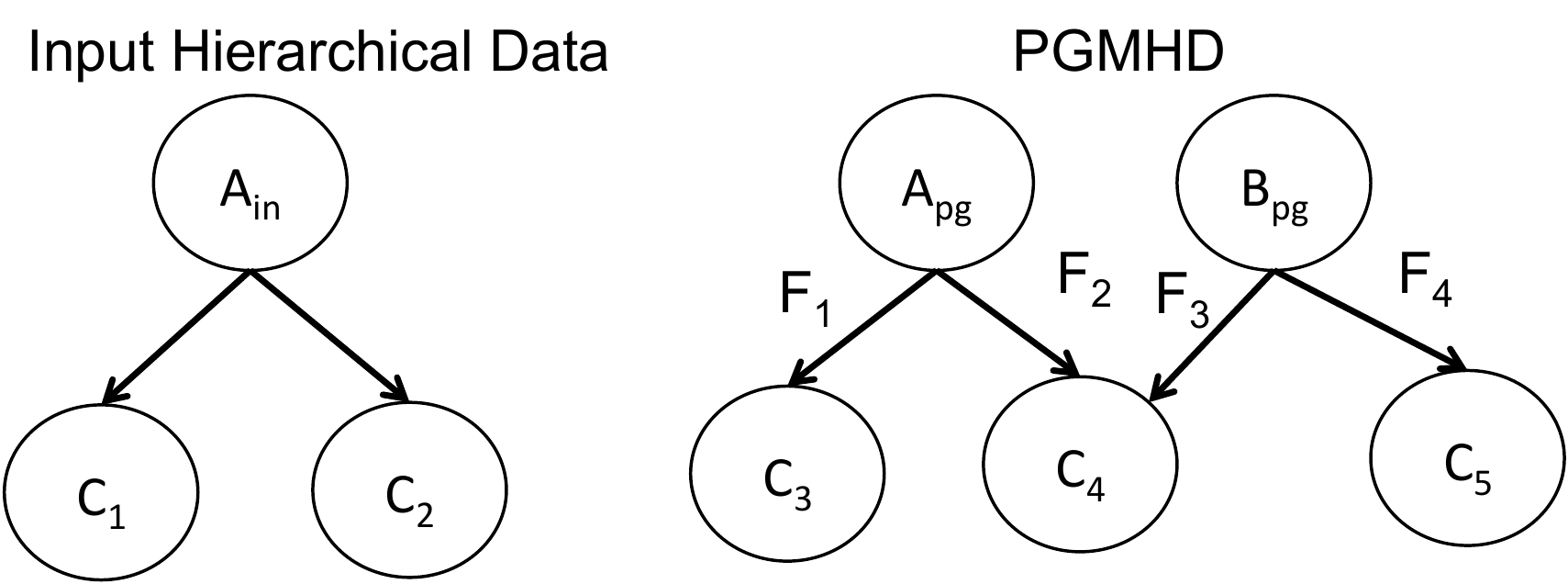}
\caption{Input hierarchical data and PGMHD}
\label{fig:input_graph}
\end{figure}

\begin{algorithm}
\label{alg:learn}
\KwData{Input Hierarchical Data}
\KwResult{PGMHD Instance}

\Begin{
$currentLevel = 0$\\
\While {$currentLevel < maxInputLevel$}{
\ForEach {$inputNode \in pgmhd(currentLevel)$}{
 \eIf {$inputNode$ exists in $PGMHD$}{
 get $pgmhdNode$ where $pgmhdNode.data$ = $inputNode.data$\\
 $pgmhdNode.frequency += 1$
 }{
 $pgmhdNode = new node$\\
 $pgmhdNode.frequency = 1$
 }
 $childrenLevel = currentLevel+1$\\
 \ForEach {$inputChildNode \in inputNode.children$}{
 \ForEach {$pgmhdChildNode \in pgmhdNode.children$}{
 \eIf {$inputChildNode.data = pgmhdChildNode.data$}{
 $edge = edge(pgmhdNode,pgmhdChildNode)$
 $edge.frequency += 1$
 }{\eIf{$childNode \in pgmhd(childrenLevel)$}{
 $pgmhdChildNode = node$ where $node.data = childNode.data$
 $edge = createNewEdge(pgmhdNode,pgmhdChildNode)$
 $edge.frequency = 1$
 }{
 $pgmhdChildNode = new Node$
 $pgmhdChildNode.data = child$
 $pgmhdChildNode.frequency = 1$
 $edge = createNewEdge(pgmhdNode,pgmhdChildNode)$
 $edge.frequency = 1$
 }}
 }
}
}
  $currentLevel = currentLevel+1$
}}
\caption{Learning Algorithm for PGMHD. $currentLevel$ represents the current level in the input hierarchical data, we start with level$_0$. $maxInputLevel$ is the highest level in the input hierarchical data. In Figure \ref{fig:input_graph} $A_{in}$ is an $inputNode$, while $A_{pg}$ is the $pgmhdNode$. $C_1$ is $inputChildNode$. $C_3$ is $PgmhdChildNode$. $F_1$ is $edge.Frequency$. $C_1,C_2 \in inputParentNode.children$. $C_3,C_4,C_5 \in PgmhdParentNode.children$}
\end{algorithm}

\subsection*{PGMHD an extension to NB}
PGMHD extends NB in different directions to improve its scalability and ability to handle massive hierarchical data as follows:
\begin{enumerate}
\item It enables multi-label classification.
\item It enables multi-level representation of the predictive features.
\item It enables lazy classification.
\end{enumerate}
  The first dimension PGMHD extends is the multi-label classification. Our model allows more than one class to be in the root level of the classifier where any instance can be classified to more than one class. The second dimension of this extension is the multi-level classification which allows the classifier to represent the predictive features in $m$ levels instead of only 2 levels as in the regular NB. This extension allows the hierarchical modeling to preserve the structure of the data, which our experiments show is important for improving the quality of the classification. The last dimension of this extension is lazy classification against the eager NB. PGMHD is considered a lazy classifier since the calculation of the classification score of a new instance is all done during the classification process, unlike NB where all the CPT are pre-calculated and stored. This extension makes the PGMHD suitable for progressive learning, which can be very important for scalability.
\section*{Experiments and Results}
\label{sec:Exp} 
\begin{figure}
\centering
\includegraphics[scale=0.5]{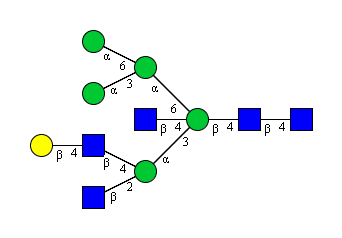}
\caption{Glycan structure in CFG format. The circles and squares represent the monosaccharides which are the building blocks of a glycan while the lines are the linkages between them}
\label{glycan}
\end{figure}

Glycans (Figure \ref{glycan}) are the third major class of biological macro-molecules besides nucleic acids and proteins \cite{9}. Glycomics refers to the scientific attempts to characterize and study glycans, as defined in \cite{9} or an integrated systems approach to study structure-function relationships of glycans as defined in \cite{10}.

Mass spectrometry (MS) is an analytical technique used to identify the composition of a sample \cite{ma2010challenges}. Although (MS) has become the major analytical technique for glycans, no general method has been developed for the automated identification of glycan structures using MS and tandem MS data. MS$^n$ refers to the sequence of MS selection with some form of fragmentation, it is also called tandem MS. The relative ease of peptide identification using tandem MS is mainly due to the linear structure of peptides and the availability of reliable peptide sequence databases. In proteomic analyses, a mostly complete series of high abundance fragment ions is often observed. In such tandem mass spectra, the mass of each amino acid in the sequence corresponds to the mass difference between two high-abundance peaks, allowing the amino acid sequence to be deduced. In glycomics MS data, ion series are disrupted by the branched nature of the molecule, significantly complicating the extraction of sequence information. In addition, groups of isomeric monosaccharides commonly share the same mass, making it impossible to distinguish them by MS alone. Databases for glycans exist but are limited, minimally curated, and suffer badly from pollution from glycan structures that are not produced in nature or are irrelevant to the organism of study.
PGMHD attempts to employ machine learning techniques (mainly probabilistic-based multi-label classification) to find a solution for the automated identification of glycans using MS data.

\begin{figure}
\centering
\includegraphics[scale=0.85]{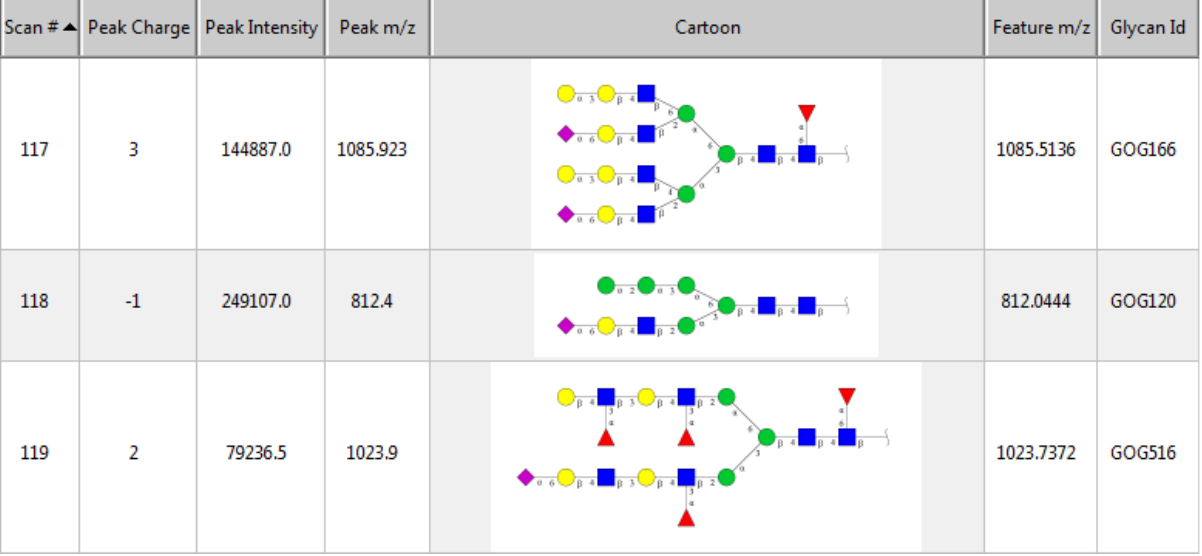}
\caption{MS$^1$ annotation using GELATO. Scan is the ID number of the scan
in the MS file, peak charge is the charge state of that peak in the MS file,
peak intensity represents the abundance of an ion at that peak, peak m/z is
the mass over charge of the given peak, cartoon is the annotation of that peak
(glycan) in CFG format, feature m/z is the mass over charge for the glycan,
and glycanID is the ID of the glycan in the Glycan Ontology (GlycO).}
\label{MS1}
\end{figure}

\begin{figure}
\centering
\includegraphics[scale=0.85]{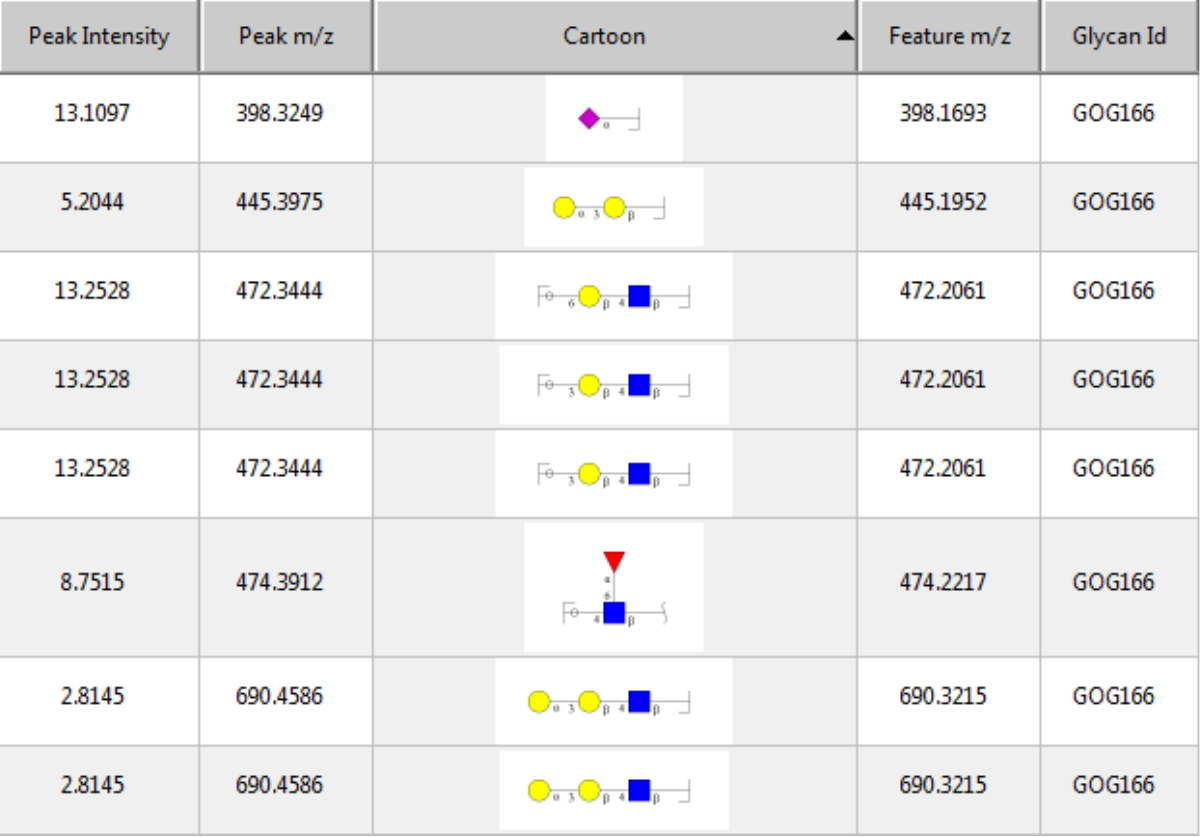}
\caption{Fragments of a selected glycan at the MS$^2$ level. Each ion observed in MS$^1$ is selected and fragmented in MS$^2$ to generate smaller ions, which can be used to identify the glycan structure that most appropriately annotates the MS$^1$ ion. Theoretical fragments of the glycan structure that had been used to annotate the MS$^1$ spectrum are used to annotate the corresponding MS$^2$ spectrum.}
\label{MS2_1}
\end{figure}

We recently implemented the Glycan Elucidation and Annotation Tool (GELATO), which is a semi-automated MS annotation tool for glycomics integrated within our MS data processing framework called GRITS (http://www.grits-toolbox.org/). Figures \ref{MS1}, and \ref{MS2_1} show screen shots from GELATO for annotated spectra. Figure \ref{MS1} shows the MS profile level and Figure \ref{MS2_1} shows the annotation of MS$^2$ peaks using fragments of a selected candidate glycan for annotation of the MS$^1$ data. The output GELATO represents all the possible annotations to the given spectra. The user may select a subset of those possible annotations as the correct ones, but then he/she needs a smarter tool that can learn the correct selection and eliminate the incorrect ones in the future. PGMHD is successfully applied for that purpose as we show in this section. 

\begin{figure}
\centering
\includegraphics[scale=0.5]{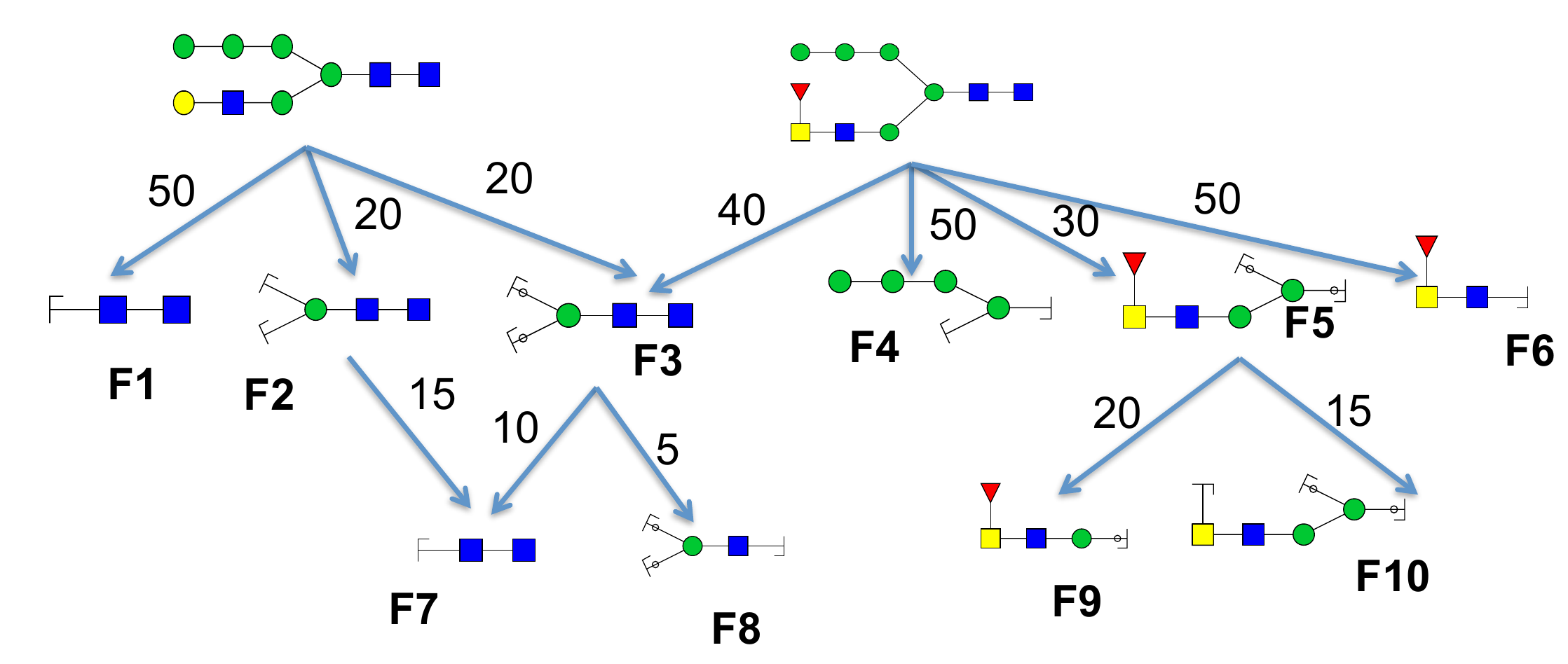}
\caption{PGMHD representing MS annotations. The root nodes are the glycans that annotate the peaks at MS$^1$ level, while the level 2 and 3 nodes are the glycan fragments that annotate the peaks at MS$^2$ and MS$^3$ level respectively and the edges represent dependency associating the glycans with their MS$^2$ fragments and MS$^2$ with their MS$^3$ fragments.}
\label{pgmhd_ms}
\end{figure}

To represent the MS data annotation using PGMHD, each annotation of $MS^1$ data (which is a glycan) is represented as a node in the top-layer of PGMHD. All the fragments generated by that glycan and used to annotate peaks in $MS^2$ are represented by nodes in the lower layer and connected by edges with the parent node in the upper layer, and this pattern can be extended until $MS^n$. Each fragment at level $MS^i$ is represented by a node in layer $L_{i-1}$ and connected by an edge with its parent node at layer $L_{i-2}$. The edge's weight represents the co-occurrence frequency between a child and a parent, and storing frequencies rather than probabilities facilitates progressive learning. Figure \ref{pgmhd_ms} shows the PGMHD for MS data with three levels ($MS^1$, $MS^2$, and $MS^3$) in these figures. As shown in the model, three layers are created: one for the MS$^1$ level, a second one for the MS$^2$ level, and a third for $MS^3$. Several different nodes at the MS$^1$ level can be annotated with the same fragment ion at the MS$^2$ level, so MS$^2$ nodes can have several parents. The frequency values are shown on the edges.  

\begin{figure}
\centering
\includegraphics[scale=0.5]{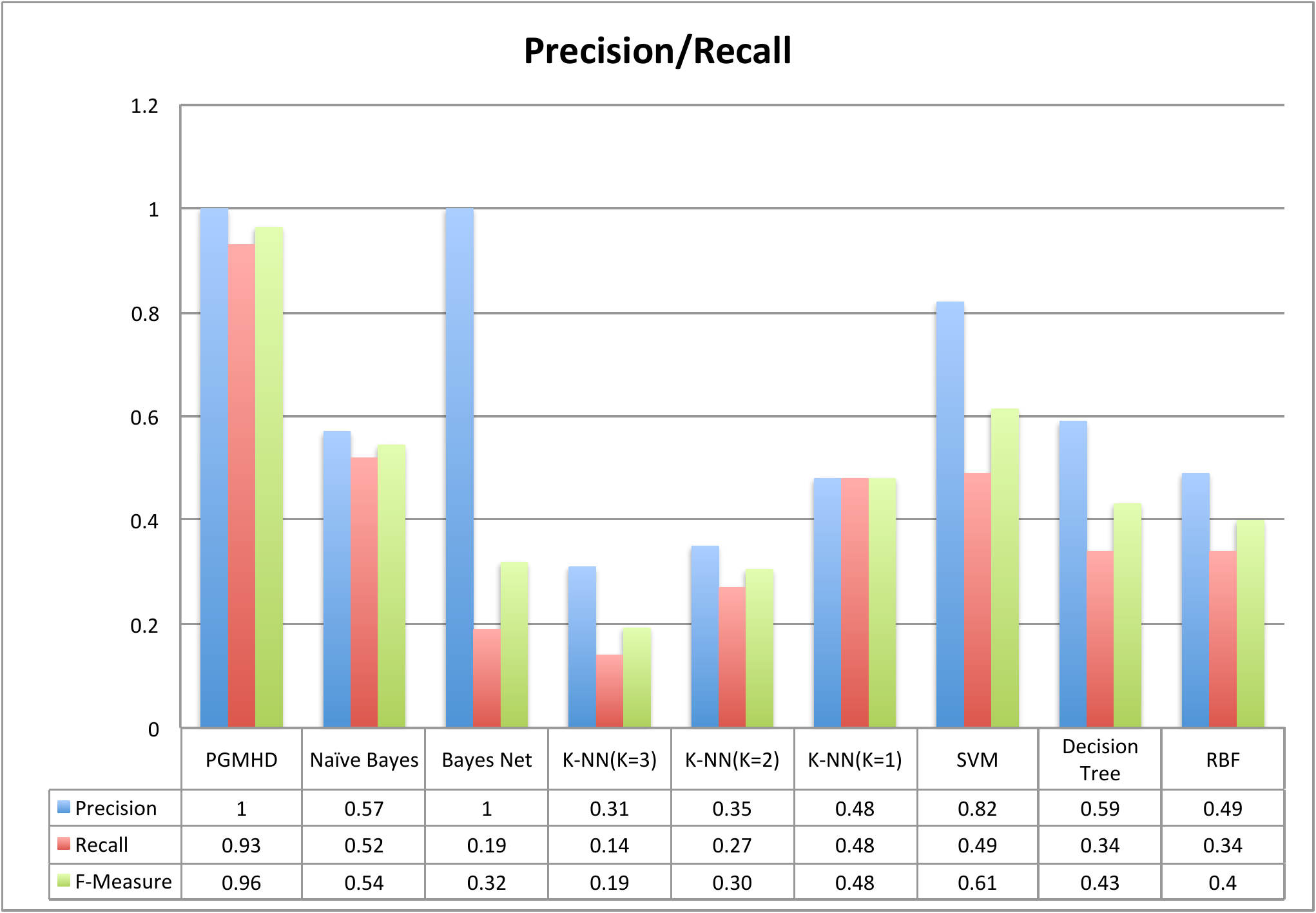}
\caption{Precision and Recall after the Multi-label classification. PGMHD was applied with the m-estimate where $m=1$ and $p=0.1$}
\label{Prec_Rec_MS}
\end{figure}

\begin{figure}
\centering
\includegraphics[scale=0.5]{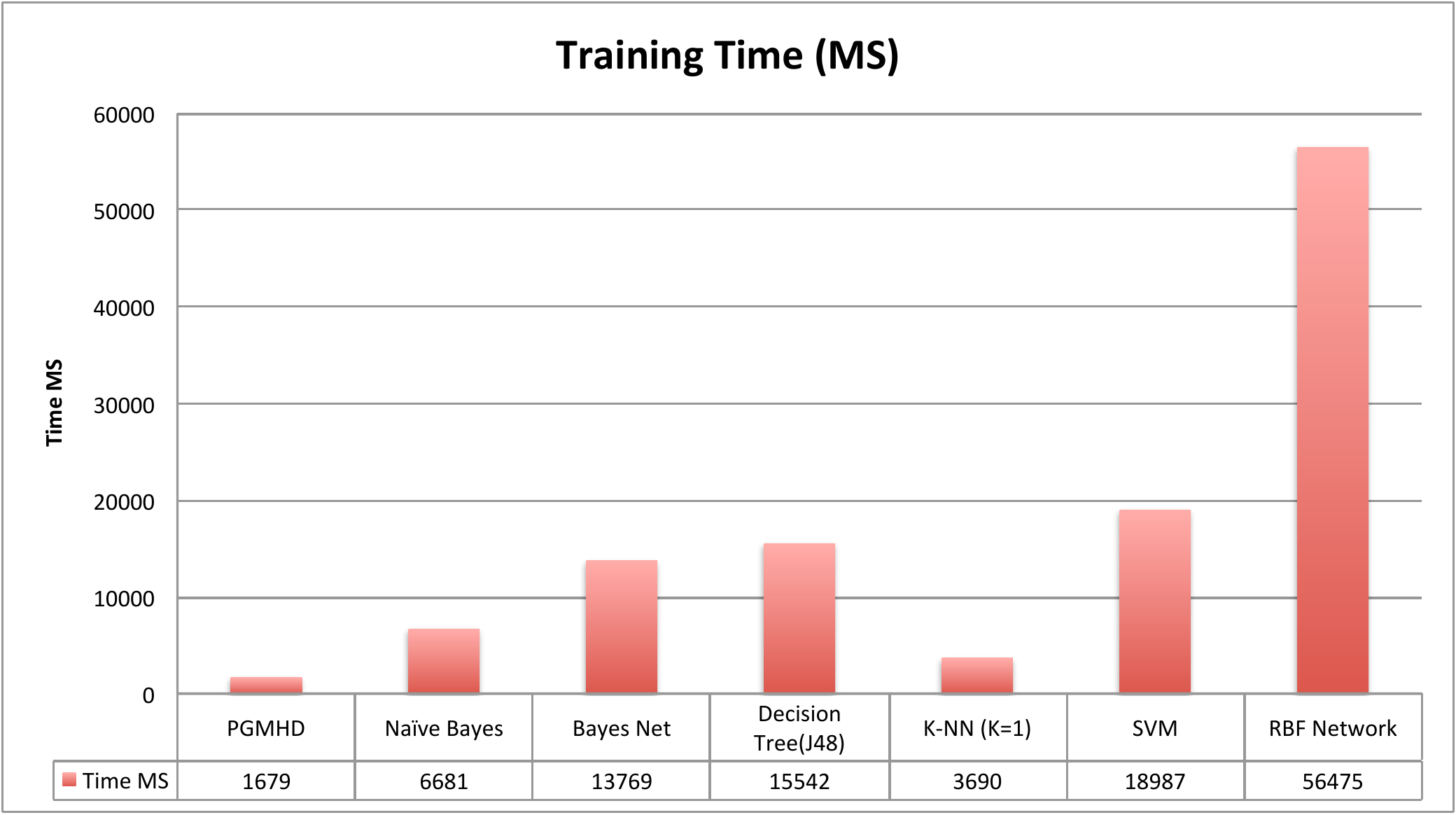}
\caption{Training time for different classifiers. Lazy classifiers (PGMHD, and K-NN) are much faster in the training phase due to the fact that no complicated calculation is required.}
\label{Training}
\end{figure}

\begin{figure}
\centering
\includegraphics[scale=0.5]{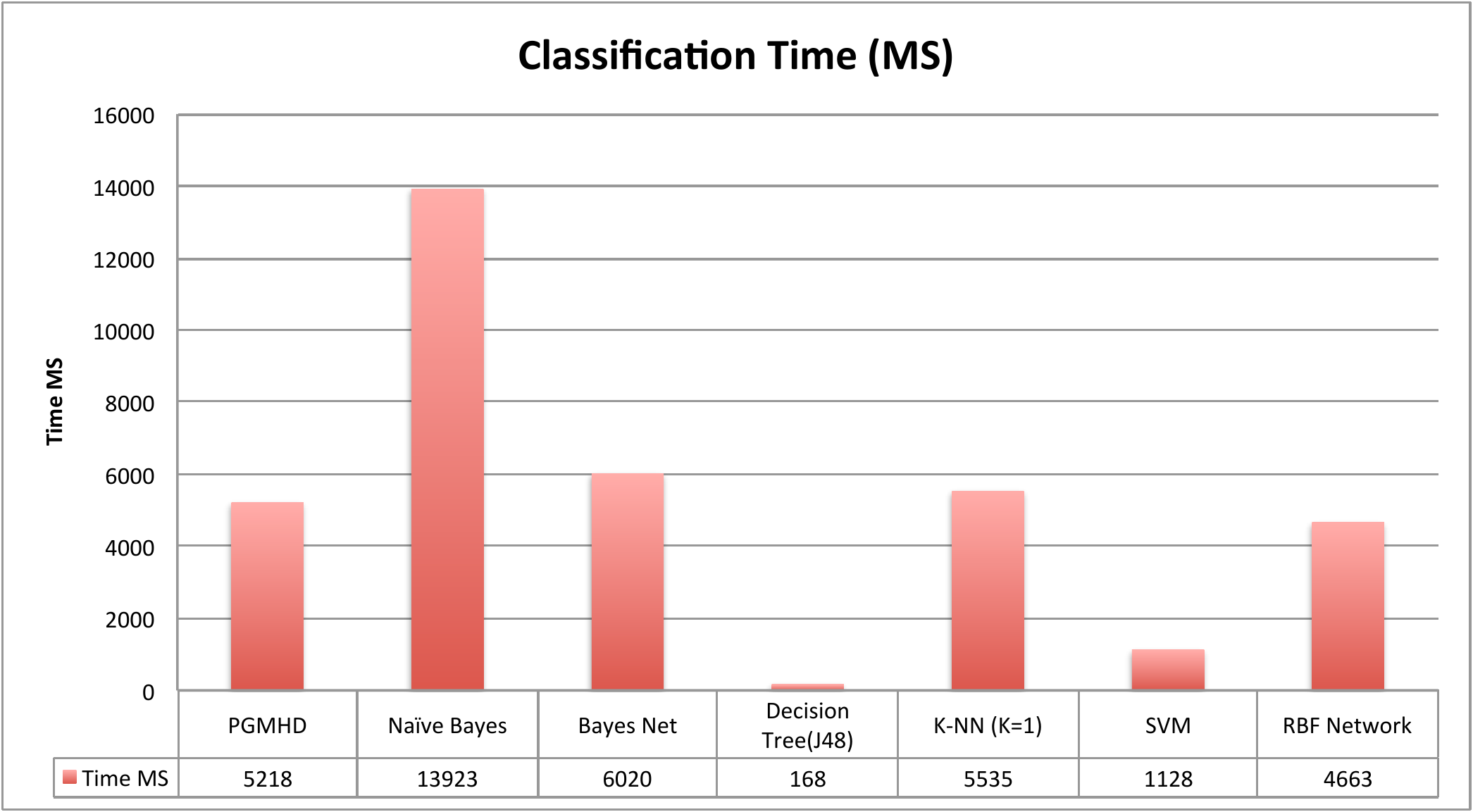}
\caption{Classification time for different classifiers. The eager classifiers (Decision Tree, SVM, and RBF) are faster than the lazy ones due to the fact that the complicated computations are done during the training phase, which causes the classification time to be faster. Naive Bayes and Bayesian Networks did not do well due to the multi-label classification for which they are not suitable.}
\label{Classification}
\end{figure}

\begin{figure}
\centering
\includegraphics[scale=0.5]{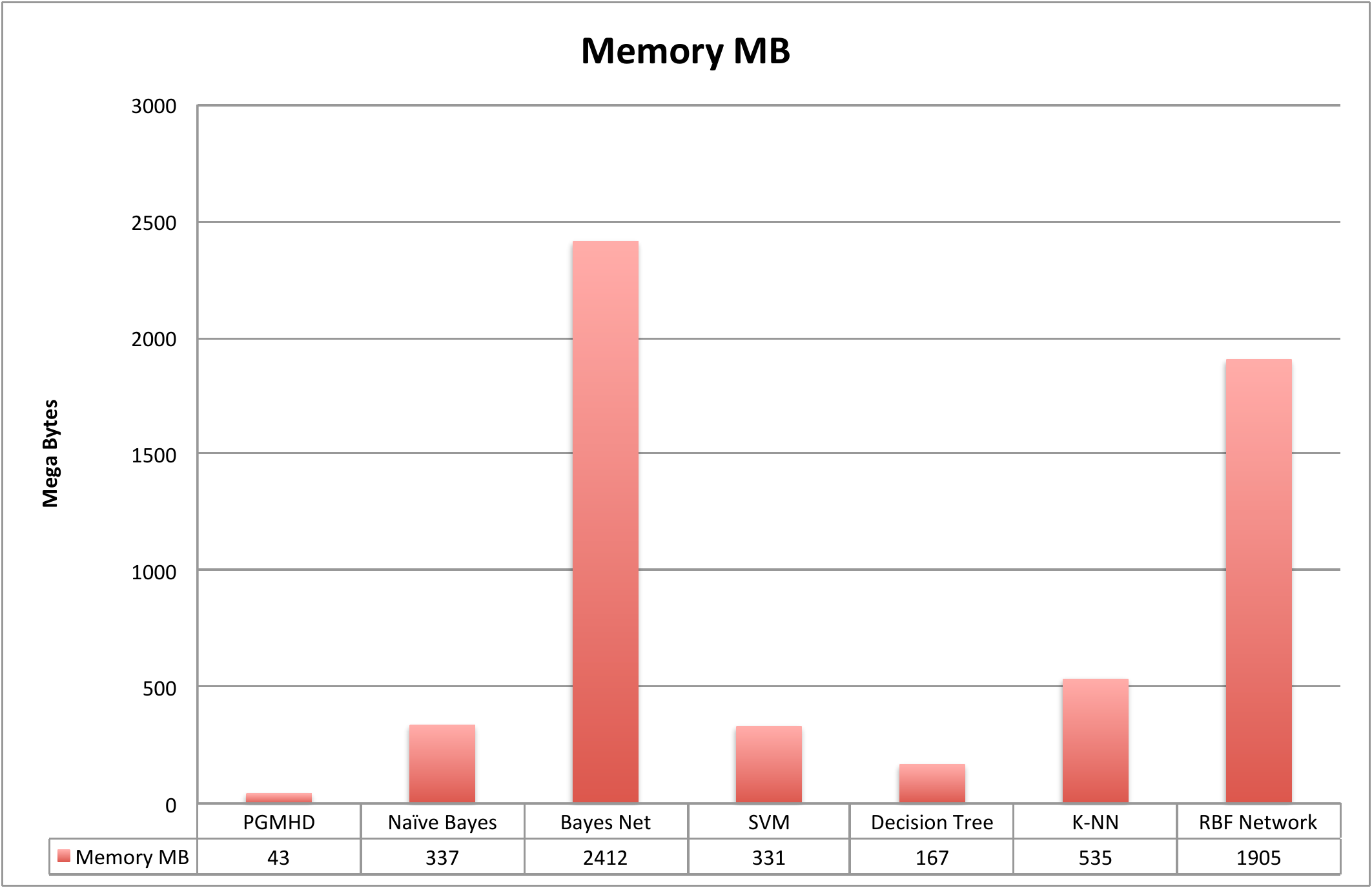}
\caption{Memory usage by each model in MB for a dataset of 1779 instances annotated by 468 glycans (classes).}
\label{Memory}
\end{figure}

\begin{figure}
\centering
\includegraphics[scale=0.5]{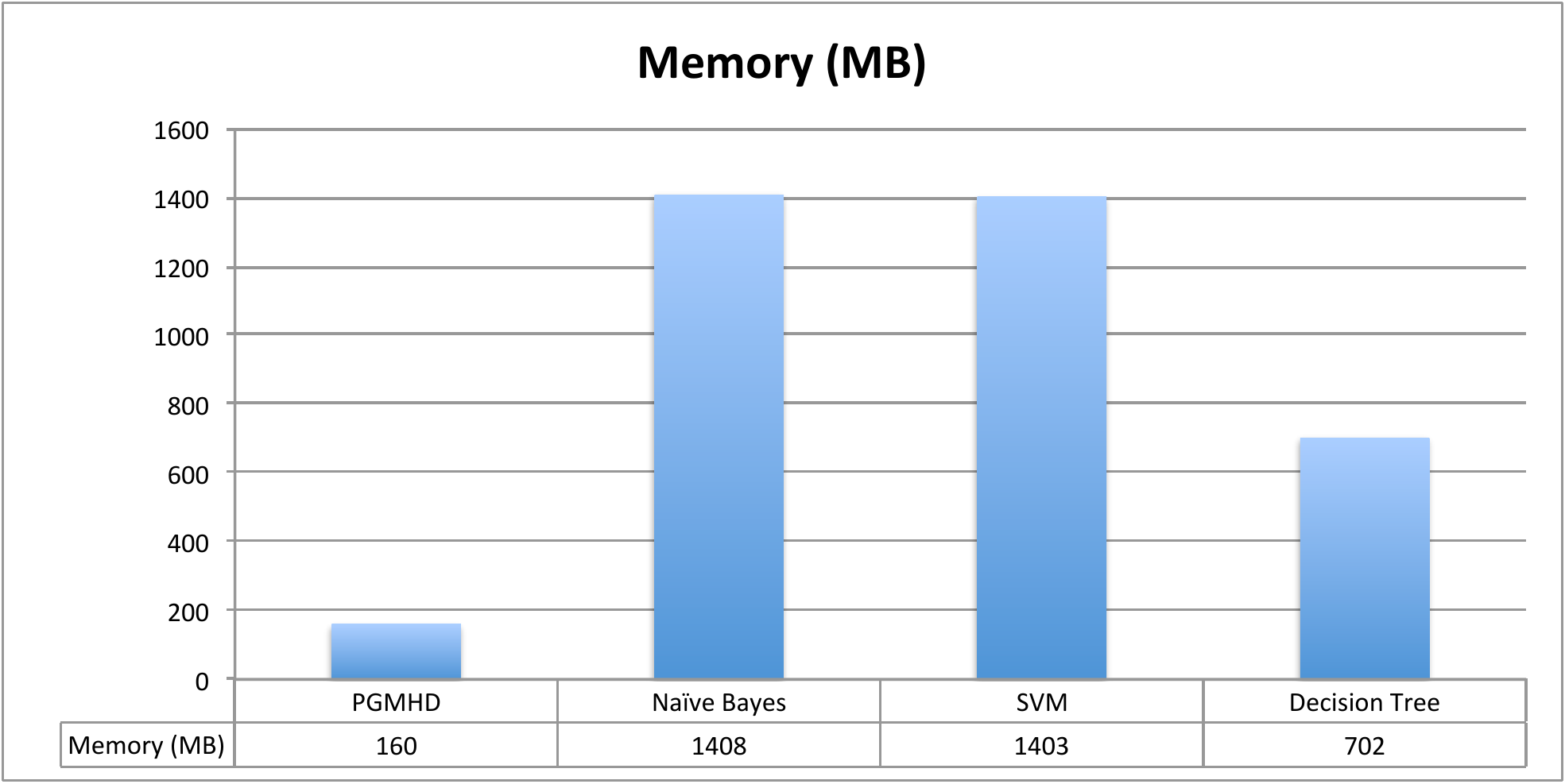}
\caption{Memory usage by each model in MB for a training dataset of 6776 instances, 1640 features, and annotated by 1340 glycans (classes).}
\label{exp2}
\end{figure}

We annotated 3314 MS scans of banceriatic cancer samples using GELATO. Then an expert manually approved 1990 scan annotations which we used to train and test PGMHD. We split this data set into training data and test data sets. The size of the training data is 1779 scans and 121 scans for testing. We trained PGMHD and compared it against leading classifiers including Naive Bayes \cite{rish2001empirical}, SVM \cite{cortes1995support}, Decision Tree \cite{quinlan1987simplifying}, K-NN \cite{altman1992introduction}, Neural Network \cite{werbos1988generalization}, Radial Basis Function network (RBF Network) \cite{poggio1989theory} and Bayesian Network \cite{pearl1985bayesian} from Weka \cite{hall2009weka}. Then we provide the test list to each classifier to predict the best glycan that annotates the scans in the test set. We also used Mulan \cite{tsoumakas2011mulan}, which is a Java library that extends Weka classifiers to handle multi-label classification problems. Also, we applied the m-estimate as a probability estimation technique To help PGMHD overcome the common problem for any Bayesian model which is the zero-frequency problem \cite{jiang2007scaling}. Figure \ref{Prec_Rec_MS} shows the precision and recall for the different classifiers compared to PGMHD after we used Mulan for multi-label classification and m-estimate for PGMHD. Another important aspect in our experiment besides accuracy is the scalability. In order to measure the scalability of PGMHD compared to the other classifiers we measure the space and time complexity. Figure \ref{Training} shows how PGMHD was the fastest model in the training phase, however it was not the best in the classification time as shown in Figure \ref{Classification}, though it did get the third best time. Most important is the space complexity used by each model which is shown in Figure \ref{Memory}. The memory usage which is the most important aspect in the scalability shows that PGMHD is much better than all the other classifiers especially the bayesian ones. Due to the difficulty of getting more manually curated MS annotations dataset for testing the scalability of our model in comparison to other machine learning models, we synthesized a dataset using GELATO. To synthesize a dataset with a massive number of MS annotations we used GELATO to generate all the possible annotations for the MS experiments which were manually curated before. We assume that all the generated annotations by GELATO are valid and correct annotations. Our focus in this part of the experiment is the scalability not the accuracy since the accuracy was already tested using the manually curated dataset. The new dataset includes 6776 instances for training and 392 instances for testing. The number of features is 2952 while the number of classes is 1340. As a result of this extension in the training data, the Baysian Network classifer, K-NN, and RBF ran out of memory which means they can not handle this dataset in 4 GB of main memory. On the other hand PGMHD used only 160 MB to represent this dataset in memory. Figure \ref{exp2} shows the memory usage of the models which scale successfully to handle the new dataset.


\subsection*{Semantically Related Keywords in Search Logs}

Semantic similarity is a metric that is defined over documents or terms in which the distance between them reflects the likeness of their meaning ~\cite{harispe2013semantic}, and it is widely used in Natural Language Processing (NLP) and Information Retrieval (IR) \cite{mihalcea2006corpus}. Generally, there are two major techniques used to compute semantic similarity: one is computed using a semantic network (Knowledge-based approach) \cite{budanitsky2001semantic}, and the other is based on computing the relatedness of terms within a large corpus of text (corpus-based approach) \cite{mihalcea2006corpus}. The major techniques classified as corpus-based approaches are Pointwise Mutual Information (PMI) \cite{bouma2009normalized} and Latent Semantic Analysis (LSA) \cite{dumais2004latent}, though PMI outperforms LSA on mining the web for synonyms \cite{turney2001mining}. A group of Google researchers proposed two efficient models which can discover semantic word similarities \cite{mikolov2013efficient}. The two novel models are the following:
\begin{enumerate}
\item Continuous Bag-of-Words model
\item Continuous Skip-gram model
\end{enumerate}
These models aim to use large scale Neural Network to learn word vectors. The two models have restrictions that make them not suitable in our usecase. The first restriction is that both models require words and context in which those words are used. In their experiments, the authors built vectors of at least 50 words around the given word (words before and after the given word from the text in which that word was used). One more restriction is that they allow only single token words to be processed (no phrases). In our case, the two models are not applicable since the searches conducted by the users usually contains a single phrase with no context or other words surrounding it. Also, we care about phrases as opposed to single words, since small phrases are most commonly used in our search engine. For example "$Java\ Developer$" should be considered as a single phrase when we discover the semantically related phrases. In our experiment, we discovered high quality semantic relationships using a data set of 1.6 billion search logs (search keywords used to search for jobs on CareerBuilder.com). PGMHD completed this task in 45 minutes.
\subsubsection*{Motivation}
We would like to create a language-independent algorithm for modeling semantic relationships between search phrases that provides output in a human-understandable format. It is important that the person searching can be assisted by an augmented query without us creating a black-box system in which that person is unable to understand and adjust the query augmentation.
CareerBuilder\footnote{http://www.careerbuilder.com/}
operates job boards in many countries and receives tens of millions of search queries every day.  Given the tremendous volume of search data in our logs, we would like to discover the latent semantic relationships between search terms and phrases for different region-specific websites using a novel technique that avoids the need to use natural language processing (NLP).  We wish to avoid NLP in order to make it possible to apply the same technique to different websites supporting many languages without having to change the algorithms or the libraries per-language.

It is tempting to suggest using a synonym dictionary since the problem sounds like finding synonyms, but the problem here is more complicated than finding synonyms since the search terms or phrases on our site are often job titles, skills, and company names which are not, in most cases, regular words from any dictionary. For example if a user searches for $"java\ developer"$, we would not find any synonyms for this phrase in a dictionary.  Another user may search for $"hadoop"$ which is also not a word that would be found in a typical English dictionary.
\subsubsection*{Probabilistic Semantic Similarity Scoring using PGMHD}
\begin{figure}[h!]
\centering
\includegraphics[scale=0.5]{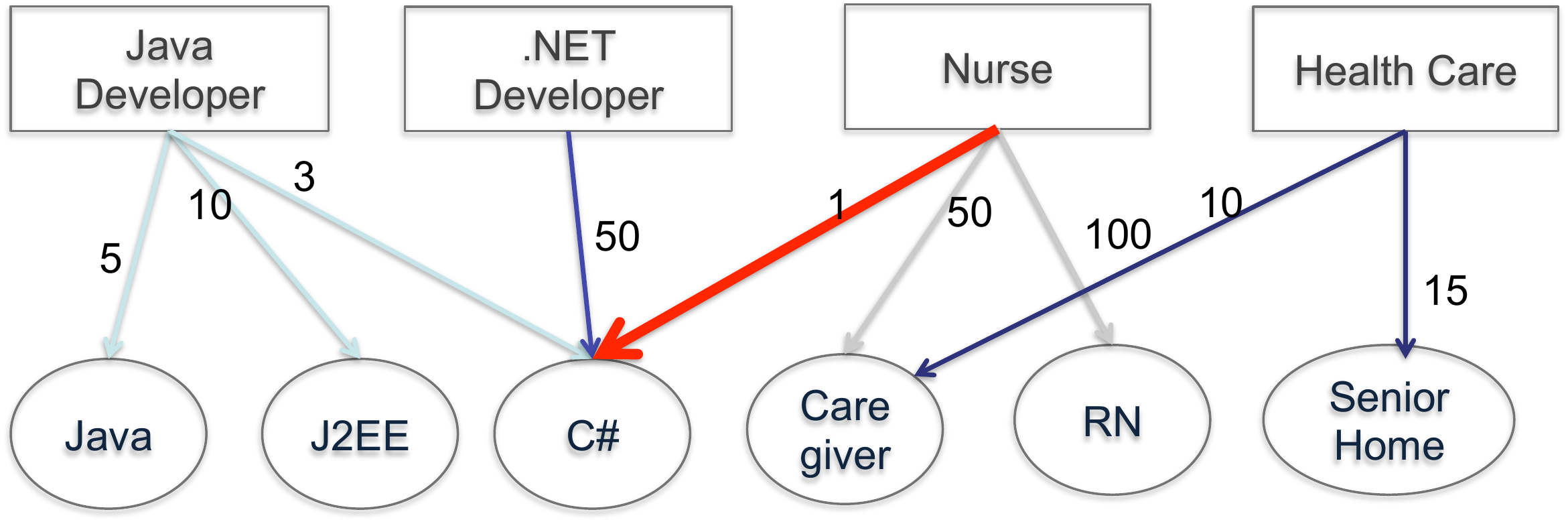}
\caption{PGMHD Representing Job Search Keywords}
\label{pgmhdJobs}
\end{figure}

We applied the proposed PGMHD model to discover the semantically related search terms by measuring probabilistic-based semantic similarity between those search terms.
Given the search logs for all the users and the users' classifications as shown in Table \ref{input_pgmhd}, PGMHD can represent this kind of data by placing the classes of the users as root nodes and placing the search terms for all the users in the second level as children nodes. Then, an edge will be formed linking each search term back to the class of the user who searched for it. The frequency of each search term (how many users search for it) will be stored in the node of that term, while the frequency of a specific search term searched for by users of a specific class (how many users belonging to that class searched for the given term) will be stored in the edge between the class and the term. The frequency of the root node is the summation of the frequencies on the edges that connect that root node with its children (Figure \ref{pgmhdJobs}).

\begin{table}[!t]
\renewcommand{\arraystretch}{1.3}
\centering
\caption{Input data to PGMHD over hadoop}
\label{input_pgmhd}
\begin{tabular}{c|c|c}
\hline
\bfseries UserID & \bfseries Classification &\bfseries Search Terms \\
\hline
\hline
user1 & Java Developer & Java, Java Developer, C$\#$, Software Engineer\\
\hline
user2 & Nurse & RN, Rigistered Nurse, Health Care\\
\hline
user3 & .NET Developer & C$\#$, ASP, VB, Software Engineer, SE\\
\hline
user4 & Java Developer & Java, JEE, Struts, Software Engineer, SE\\
\hline
user5 & Health Care & Health Care Rep, HealthCare\\
\hline
\end{tabular}
\end{table}

\subsubsection*{Distributed PGMHD}
\begin{figure}[h!]
\centering
\includegraphics[scale=0.5]{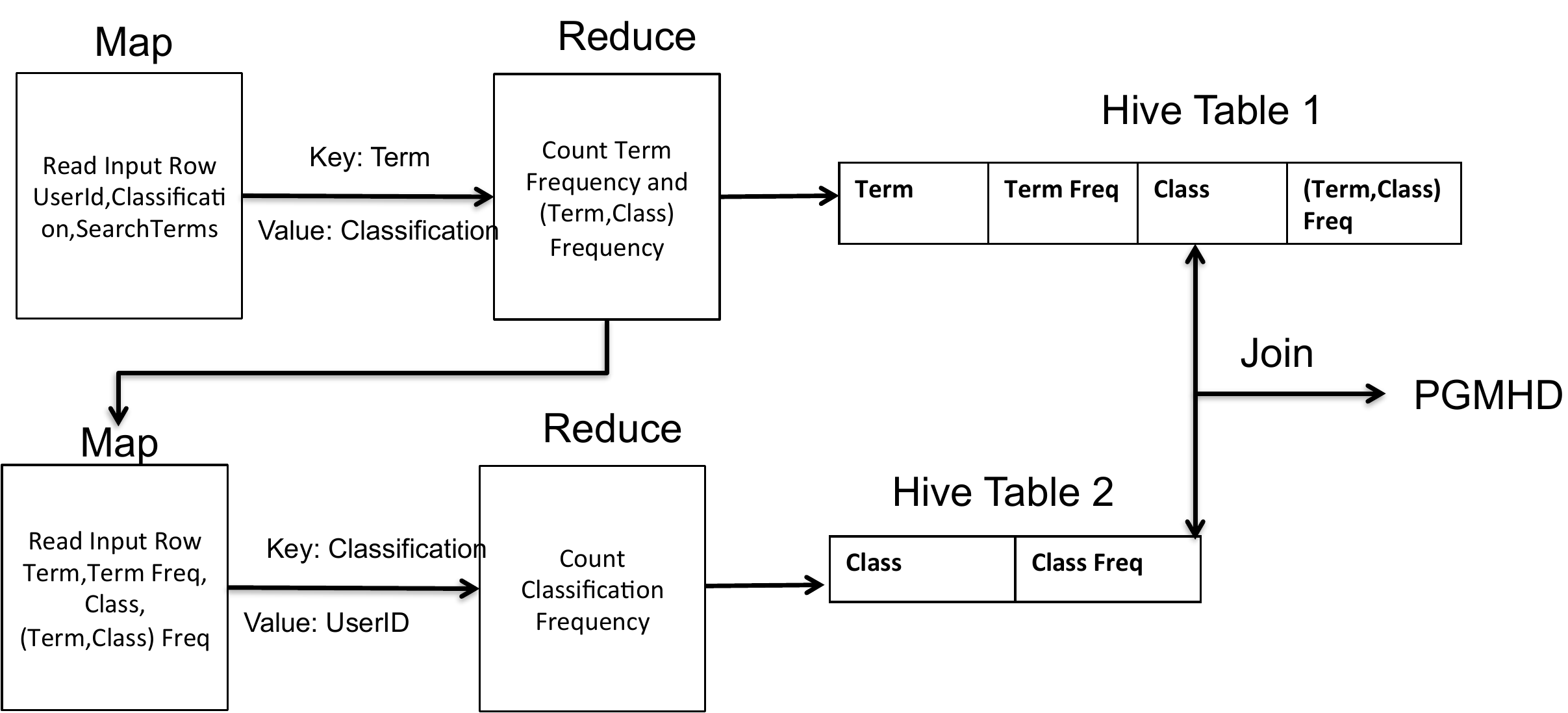}
\caption{PGMHD Over Hadoop}
\label{pgmhdHadoop}
\end{figure}

In order to process 1.6 billion search logs (each search log contains one or more keywords used by a user to search for jobs on careerbuilder.com) provided by Careerbuilder in reasonable time, we designed a distributed PGMHD using Hadoop HDFS \cite{shvachko2010hadoop}, Hadoop Map/Reduce \cite{dean2008mapreduce} and Hive \cite{thusoo2009hive}. The design of distributed PGMHD is shown in figure \ref{pgmhdHadoop}. Basically we use Hive to store the intermediate data while we are buidling and training PGMHD. Once it is trained we can then run our inquires to get an ordered list of the semantically related keywords for a specific term(s).

\subsubsection{Experiment Setup and Results}

The experiment performing latent semantic discovery among search terms using PGMHD was run on a Hadoop cluster with 69 data nodes, each having a 2.6 GHz AMD Opteron Processor with 12 to 32 cores and 32 to 128 GB RAM. Table \ref{pgmhdResults} shows sample results of 10 terms with their top 5 related terms discovered by PGMHD. To evaluate the model's accuracy, we sent the results to data analysts at CareerBuilder who reviewed 1000 random pairs of discovered related search terms and returned the list with their feedback about whether each pair of discovered related terms was "related" or "unrelated". We then calculated the accuracy (precision) of the model based upon the ratio of the number of related results to the total number of results. The results show the accuracy of the discovered semantic relationships among search terms using the PGMHD model to be 0.80.

\begin{table}[h!]
\centering
\caption{\label{pgmhdResults}PGMHD results for latent semantic discovery}
\begin{tabularx}{\textwidth}{|c|X|}
\hline
\bfseries Term &  \bfseries Related Terms\\
\hline
hadoop & big data, hadoop developer, obiee, java, python\\
\hline
registered nurse & rn registered nurse, rn, registered nurse manager,  nurse, nursing, director of nursing\\
\hline
data mining & machine learning, data scientist,  analytics, business intellegence, statistical analyst\\
\hline
solr & lucene, hadoop, java\\
\hline
software engineer & software developer, programmer, .net developer, web developer, software \\
\hline
big data & nosql, data science, machine learning, hadoop, teradata \\
\hline
realtor & realtor assistant, real estate, real estate sales, sales,
real estate agent \\
\hline
data scientist & machine learning, data analyst, data mining, analytics,
big data\\
\hline
plumbing & plumber, plumbing apprentice, plumbing maintenance,
plumbing sales, maintenance\\
\hline
agile & scrum, project manager, agile coach, pmiacp,
scrum master\\

\hline
\end{tabularx}
\end{table}

\subsection*{Discovering Semantic Ambiguity of a Keyword}
The semantic ambiguity of a keyword can be defined as the likelihood of seeing different meanings of the same keyword in different contexts \cite{jayadianti2013solving,gracia2007solving}. The techniques mentioned in the literature focuses on utilization of ontologies and dictionaries like Wordnet as described in \cite{jayadianti2013solving,gracia2007solving}. Those solutions are not applicable when the keywords are from a domain like job search. In the job search domain the used keywords are typically job titles, skills, company names, etc. which are not  regular English keywords. For example, java can mean a programming language, as well as,  coffee but an English dictionary would not provide both of those meanings.

PGMHD is applied successfully to discover the semantic ambiguity of a keyword. About 1.6 billion search logs used in this experiment. The search keywords extracted from those 1.6 billion logs were used to train PGMHD, which was then used to calculate the normalized PMI score for each term with all of its parents. The initial results of this use case are promising, though work to improve the implementation are is still ongoing. We plan to publish a separate paper about this use case of PGMHD soon.

Table \ref{DisAmbResults} shows sample results of the discovered semantically ambiguous terms using PGMHD.

\begin{table}[h!]
\centering
\caption{\label{DisAmbResults}PGMHD results for semantic ambiguity discovery. The first column shows the keyword, while the second column shows the related keywords of each possible meaning separated by horizontal line}
\begin{tabular}{>{\centering}m{.09\textwidth}|>{\raggedright}p{.91\textwidth}}
\hline 
\textbf{term} & \textbf{related terms of meaning}\tabularnewline
\hline 
\multirow{2}{2cm}{architect} & enterprise architect, java architect, data architect, telecommute,
oracle, java, .net\tabularnewline
\cline{2-2} 
 & architectural designer, architectural drafter, autocad, autocad drafter,
designer, drafter, cad, engineer\tabularnewline
\hline 
\multirow{2}{2cm}{account} & bookkeeper, accountant, analyst, finance\tabularnewline
\cline{2-2} 
 & sales executive, account executive, insurance, account manager, outside
sales, medical sales, manager, sales\tabularnewline
\hline 
\multirow{3}{2cm}{designer} & design, print, animation, artist, illustrator, creative, graphic artist,
graphic, photoshop, video\tabularnewline
\cline{2-2} 
 & graphic, web designer, design, web design, graphic design, graphic
designer\tabularnewline
\cline{2-2} 
 & design, drafter, cad designer, draftsman, autocad, mechanical designer,
proe, drafter drafting designer autocad, structural designer, revit\tabularnewline
\hline 
\end{tabular}
\end{table}

\section*{Conclusions and Future Work}
Probabilistic graphical models are very important in many modern applications such as data mining and data analytics. The major issue with existing probabilistic graphical models is their scalability to handle large data sets, making this a very important area for research, especially given the tremendous modern focus on big data due to the number of data points produced by modern computer systems and sensors. PGMHD is a probabilistic graphical model that attempts to solve the scalability problems with existing models in scenarios where massive hierarchical data is present. PGMHD is designed to fit hierarchical data sets of any size, regardless of the domain in which the data belongs. PGMHD can represent the hierarchical data with any number of levels, it can handle multi-label classification, and it is suitable for progressive learning since it is considered to be a lazy classifier. In this paper we present three experiments from different domains: one being the automated tagging of high-throughput mass spectrometry data in bioinformatics, the other being latent semantic discovery using search logs from the largest job board in the U.S, and the last one being identification of semantically ambiguous keywords. The three use cases in which we tested PGMHD show that this model is robust and can scale from a few thousand entries to billions of entries, and that it can also run on a single computer (for smaller data sets), as well as in a parallelized fashion on a large cluster of servers (69 were used in our experiment). PGMHD is used in production at CareerBuilder.com for discovery of semantically related keywords and semantically ambiguous keywords. The work on discovering semantically ambiguous keywords is ongoing, and we plan to publish a separate paper about it. We plan to compare machine learning algorithms implemented in Apache Spark  with PGMHD.

%


\begin{backmatter}

\section*{Competing interests}
  The authors declare that they have no competing interests.

\section*{Author's contributions}
KA carried out the design, implementation, and experiments related to MS annotation, Discovering semantically related keywords, and discovering semantically ambiguous keywords. MK participated in the design, implementation, and experiments related to discovering semantically related keywords, and discovering semantically ambiguous keywords. CO participated in the design, implementation, and experiments related to discovering semantically related keywords, and discovering semantically ambiguous keywords. TG participated in the design, and experiments related to discovering semantically related keywords, and discovering semantically ambiguous keywords. RR, WY, and MP participated in the design and validation of the MS annotation experiments. JM, KR, KK, and WY they all contributed to writing this manuscript and validating the model as well as the results.  All authors read and approved the final manuscript.

\section*{Acknowledgements}
The authors would like to thank David Crandall from Indiana University for providing very helpful comments and suggestions to improve this paper. We also would like to thank Kiyoko Aoki Kinoshita from Soka University for the valuable discussions and suggestions to improve this model. Deep thanks to the search team, the big data team, and the data science team at CareerBuilder.com for their support while implementing and test this model over their Hadoop cluster.


\bibliographystyle{bmc-mathphys} 
\bibliography{bmc_article}      




\end{backmatter}
\end{document}